%% file: arxiv.tex
\documentclass[10pt,twocolumn,letterpaper]{article}

\usepackage[pagenumbers]{cvpr} % To force page numbers, e.g. for an arXiv version
\input{tex/preamble}

\definecolor{cvprblue}{rgb}{0.21,0.49,0.74}
\usepackage[pagebackref,breaklinks,colorlinks,citecolor=cvprblue]{hyperref}
\usepackage{bm}

\def\paperID{9442} % *** Enter the Paper ID here
\def\confName{CVPR}
\def\confYear{2024}

\title{ProxyCap: Real-time Monocular Full-body Capture in World Space \\ via Human-Centric Proxy-to-Motion Learning}

\author{Yuxiang Zhang$^1$, Hongwen Zhang$^2$, Liangxiao Hu$^3$, Jiajun Zhang$^4$, Hongwei Yi$^5$,\\ Shengping Zhang$^3$, Yebin Liu$^1$\\
$^1$ Tsinghua University 
$^2$ Beijing Normal University
$^3$ Harbin Institute of Technology\\
$^4$ Beijing University of Posts and Telecommunications
$^5$ Max Planck Institute for Intelligent Systems
}

\begin{document}

\twocolumn[{%
\renewcommand\twocolumn[1][]{#1}%
\maketitle
\begin{center}
    \centering
    \captionsetup{type=figure}
    \vspace{-10mm}
    \includegraphics[width=\textwidth]{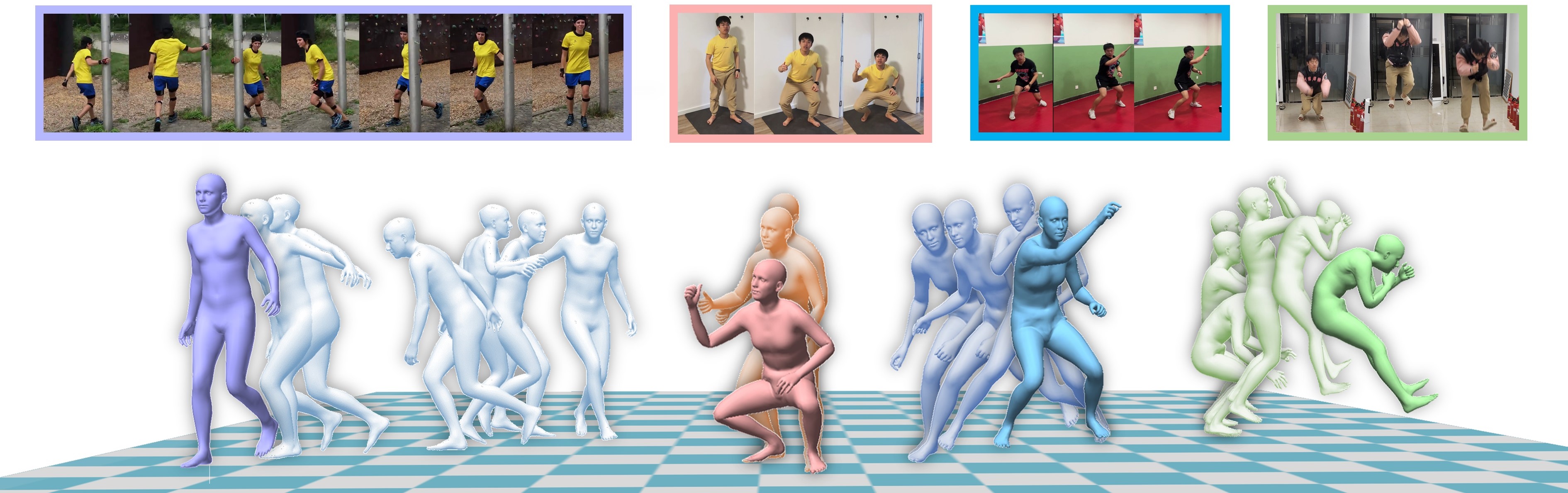}
    \vspace{-6mm}
    \captionof{figure}{The proposed method, ProxyCap, is a real-time monocular full-body capture solution to produce accurate human motions with plausible foot-ground contact in world space. }
    \vspace{-4mm}
\end{center}%
}]

\input{tex/0-abstract}

\input{tex/1-introduction.tex}

\input{tex/2-relatedwork.tex}
\input{tex/3-data.tex}

\input{tex/4-method.tex}

\input{tex/5-experiments.tex}
\input{tex/6-conclusion.tex}

\clearpage
{\Large \noindent\textbf{Supplementary Material}\\}
\input{tex/7-supp}
{
    \small
    \bibliographystyle{ieeenat_fullname}
    \bibliography{main}
}


\end{document}

%% file: tex/preamble.tex
\usepackage{booktabs} % For formal tables
\usepackage{times}
\usepackage{epsfig}
\usepackage{graphicx}
\usepackage{amsmath}
\usepackage{amssymb}
\usepackage{enumitem}
\usepackage{caption}
\usepackage{stfloats}
\usepackage{comment}
\usepackage{booktabs}
\usepackage{multirow}
\usepackage{rotating}
\usepackage{bbding} 

\usepackage[dvipsnames]{xcolor}

%% file: tex/0-abstract.tex
%%%%%%%%% ABSTRACT

\begin{abstract}
Learning-based approaches to monocular motion capture have recently shown promising results by learning to regress in a data-driven manner. However, due to the challenges in data collection and network designs, it remains challenging for existing solutions to achieve real-time full-body capture while being accurate in world space. In this work, we introduce ProxyCap, a human-centric proxy-to-motion learning scheme to learn world-space motions from a proxy dataset of 2D skeleton sequences and 3D rotational motions. Such proxy data enables us to build a learning-based network with accurate world-space supervision while also mitigating the generalization issues. For more accurate and physically plausible predictions in world space, our network is designed to learn human motions from a human-centric perspective, which enables the understanding of the same motion captured with different camera trajectories. Moreover, a contact-aware neural motion descent module is proposed in our network so that it can be aware of foot-ground contact and motion misalignment with the proxy observations. With the proposed learning-based solution, we demonstrate the first real-time monocular full-body capture system with plausible foot-ground contact in world space even using hand-held moving cameras. Our project page is \url{https://zhangyux15.github.io/ProxyCapV2}.
% enabling real-world applications even using hand-held moving cameras.
%
\end{abstract}

%% file: tex/1-introduction.tex
\section{Introduction}

% background & Target. 
% Add Why realtime is important?
% Real-time monocular full-body motion meshes capture
Motion capture from monocular videos is an essential technology for various applications such as gaming, VR/AR, sports analysis, \etc.
% by providing more natural and immersive experiences, aiding in analysis, and providing real-time feedback.
% This task has been well-studied in recent years and has yielded promising results in the regression of full-body models from monocular images.
% Despite the progress, this task is still far from being solved due to the challenges to achieve real-time full-body capture while being accurate and physically plausible in the world space.
One ultimate goal is to achieve real-time capture while being accurate and physically plausible in world space.
Despite the recent advancements, this task is still far from being solved, especially under the settings of in-the-wild captures with hand-held moving cameras.
% However, this task is highly challenging and far from being solved.
% In the field of maker-less motion capture, require tuning of hyper-parameters while typically being time-consuming and sensitive to initialization.

% explicit implicit
Compared to optimization-based methods~\cite{sigal2008combined,guan2009estimating,bogo2016keep,lassner2017unite,huang2017towards,SMPL-X:2019,yi2022mover}, learning-based approaches~\cite{kanazawa2018end,kolotouros2019learning,zhang2021pymaf,pymafx2022} can directly regress human poses from images, significantly enhancing computational efficiency while addressing the inherent issues in optimization-based methods of initialization sensitivity and local optima entrapment.
As data-driven solutions, the performance and generalization capabilities of learning-based methodologies are heavily bounded by the accuracy and diversity of the training data. Unfortunately, existing datasets are unable to simultaneously meet these requirements.
On the one hand, datasets with sequential ground truth 3D pose annotations ~\cite{ionescu2014human3, mehta2017monocular, sigal2010humaneva,huang2022capturing} are mostly captured by marker-based or multi-view systems, which makes it hard to scale up to a satisfactory level of diversity in human appearances and backgrounds. 
On the other hand, numerous in-the-wild datasets~\cite{andriluka20142d, lin2014microsoft} excel in the richness of human and scenario diversity but they lack real-world 3D motions and most of them only provide individual images instead of videos. 
%However, most of them provide only mannual 2D landmark annotations and primarily consist of individual, non-sequential frames. 
% There are continuous efforts dedicated to generating pseudo labels~\cite{kolotouros2019learning,joo2021exemplar,moon2020neuralannot} to align with the images, they inherently lack correct global motions and temporal information. 
Recently, researchers tried to create synthetic data~\cite{patel2021agora,bazavan2021hspace, Black_CVPR_2023} by rendering human avatars with controllable cameras, but it remains difficult to bridge domain gaps between the real-world images and the rendered ones, and is too expensive to scale up.

% We believe that creating virtual data is a worth exploring way to address current challenges,

% but directly rendering motion videos may be thankless in practice, for the distinct domain gap between rendered images to real data and prohibitively expensive cost to scale up.
% (silhouettes~\cite{pavlakos2018learning, varol2018bodynet}, 2D landmarks~\cite{pavlakos2018learning,sun2019human, tung2017self, moon2020i2l, moon2020pose2pose,song2020human,ma20233d,li2021hybrik}, segmentation~\cite{kocabas2021pare, omran2018neural, rueegg2020chained, kocabas2021pare}, and IUV~\cite{zhang2020learning, xu2019denserac}.) 
% Our insight is that 2D proxy representations.is more sparse, yet retain a rich amount of motion information.
In this paper, we follow the spirit of creating synthetic data, but turn to render 2D proxy representations instead of person images.
By using proxy representations (\ie, silhouettes~\cite{pavlakos2018learning, varol2018bodynet}, segmentations~\cite{kocabas2021pare, omran2018neural, rueegg2020chained, kocabas2021pare}, IUV~\cite{zhang2020learning, xu2019denserac} and 2D skeletons~\cite{pavlakos2018learning,sun2019human, tung2017self, moon2020i2l, moon2020pose2pose,song2020human,ma20233d,li2021hybrik}), the whole motion capture pipeline can be divided into two steps: image-to-proxy extraction and proxy-to-motion lifting. 
In the divided pipeline, the image-to-proxy extraction is pretty accurate and robust as there are plenty of annotated 2D ground truths in real-world datasets, while the proxy-to-motion step can leverage more diverse training data to mitigate the generalization issue and reduce the domain gap.
Here we adopt the 2D skeleton sequences as the proxy representation for its simplicity and high correlation with 3D motion. 
Meanwhile, we combine random virtual camera trajectories upon the existing large-scale motion sequence database, \ie, AMASS~\cite{mahmood2019amass}, to synthesize nearly infinite data for learning proxy-to-motion lifting.

Though the proposed proxy dataset can be generated at large scales, there remain two challenges for regression-based networks to learn physically plausible motions from proxy data: how to recover i)  world-space human motion under moving cameras, and ii) physically plausible motion with steady body-ground contact.
For a video captured with moving cameras, the trajectories/rotations of humans and cameras are coupled with each other, making the recovery of world-space human motion extremely difficult. 
To address this issue, the latest solutions~\cite{kocabas2024pace, ye2023slahmr} estimate the camera poses from the background using SfM~\cite{teed2021droid, schoenberger2016sfm}, and then estimate the human motion from a camera-centric perspective.
However, the SfM requires texture-rich backgrounds and may fail when the foreground moving character dominates the image.
Their post-processing optimization pipelines are also not suitable for real-time applications due to expensive computational costs. Besides, these previous solutions learn human motion in a camera-centric perspective like~\cite{kocabas2021spec}, which is actually ambiguous for the regression network.

In this paper, we would point out that one of the main challenges arises from the camera-centric settings in previous solutions.
In such a setting, the same motion sequence captured under different cameras will be represented as multiple human motion trajectories, making it difficult for the network to understand the inherent motion prior.
% Such a one-to-many mapping typically results in unstable learning of human motion and hinders the robustness of algorithms.
In contrast, we propose to learn human-centric motions to ensure consistent human motion outputs under different camera trajectories in synthetic data.
Specifically, our network learns the local translations and poses in a human coordinate system, together with the relative camera extrinsic parameters in this space. After that, we accumulate the local translations of humans in each frame to obtain the global camera trajectories.
Benefiting from the proposed proxy-to-motion dataset, we are able to synthesize different camera trajectories upon the same motion sequence to learn the human motion consistency.
In this way, our network can disentangle the human poses from the moving camera more easily via the strong motion prior without SfM.

% Furthermore,
% For physically plausible motion  with body-ground contact.
On top of human-centric motion regression, we further enhance the physical plausibility by introducing a contact-aware neural motion descent module.
% to refine the 2D alignments and for our initial motion estimations.
Specifically, our network first predicts coarse motions and then refines them iteratively based on foot-ground contact and motion misalignment with the proxy observations.
Compared with the global post-processing optimization used in previous work~\cite{yuan2021glamr, kocabas2024pace, ye2023slahmr}, our method learns the descent direction and step instead of explicit gradient back-propagation. 
We demonstrate that our method, termed ProxyCap, is more robust and significantly faster to support real-time applications.
To sum up, our contributions can be listed as follows:
\begin{itemize}[leftmargin=*]
\itemsep0em 
    \item To tackle the data scarcity issue, we adopt 2D skeleton sequences as proxy representations and generate proxy data in world space with random virtual camera trajectories.
    \item We design a network to learn motions from a human-centric perceptive, which enables our regressor to understand the consistency of human motions under different camera trajectories.
    \item We further propose a contact-aware neural descent module for more accurate and physically plausible predictions. Our network can be aware of foot-ground contact and motion misalignment with the proxy observations.
    \item We demonstrate a real-time full-body capture system with plausible body-ground contact in world space under moving cameras.
    % our solution can be trained end-to-end and run in real-time during inference. 
    % \item We generate proxy-to-motion datasets by synthesizing 2D pose sequences from accurate motions in the world space. Such proxy data enables us to build a learning-based motion capture method with accurate supervision while also alleviating the generalization issues. 
    % \item We propose a contact-aware neural motion descent module for more accurate and physically plausible motion prediction. This module is designed to be aware of the foot-ground contact and the motion misalignment with the proxy observations and refine the motion effectively.
    % \item We further achieve full-body motion capture with better body-hand compatibility by leveraging the body-hand context information in our network and incorporating the proxy-to-motion data of the body, hand, and face parts. As a learning-based solution, our network can be trained end-to-end and run in real-time during inference.
\end{itemize}

%% file: tex/2-relatedwork.tex
\section{Related Work}

Monocular motion capture has been an active research field recently.
We give a brief review of the works related to ours and refer readers to~\cite{tian2022recovering} for a more comprehensive survey.

\noindent \textbf{Motion Capture Datasets.} Existing motion capture datasets are either captured with marker-based~\cite{sigal2010humaneva, ionescu2014human3} or marker-less~\cite{peng2021neural, yu2020humbi,zhang20204d,zhang2021lightweight} systems.
Due to the requirement of markers or multi-view settings, the diversity of these datasets is limited in comparison with in-the-wild datasets.
To enrich the motion datasets, numerous efforts~\cite{kolotouros2019learning,joo2021exemplar,moon2020neuralannot,muller2021self, sengupta2020synthetic} have been dedicated to generating pseudo-ground truth labels with better alignment in the image space but do not consider the motion in world space.
On the other hand, researchers have also resorted to using synthetic data~\cite{varol2017learning,su2022virtualpose, patel2021agora} by rendering human models with controllable viewpoints and backgrounds.
However, such synthetic datasets are either too expensive to create or have large domain gaps with real-world images.

\noindent \textbf{Proxy Representations for Human Mesh Recovery.} Due to the lack of annotated data and the diversity of human appearances and backgrounds, learning accurate 3D motions from raw RGB images is challenging even for deep neural networks.
To alleviate this issue, previous approaches have exploited the different proxy representations, including silhouettes~\cite{pavlakos2018learning, varol2018bodynet}, 2D/3D landmarks~\cite{pavlakos2018learning,sun2019human, tung2017self, moon2020i2l, moon2020pose2pose,song2020human,ma20233d,li2021hybrik}, segmentation~\cite{kocabas2021pare, omran2018neural, rueegg2020chained, kocabas2021pare}, and IUV~\cite{zhang2020learning, xu2019denserac}.
% Pose2Pose~\cite{moon2020pose2pose} learns human poses from 3D heatmaps.
% PARE~\cite{kocabas2021pare} predicts body-part-guided attention masks and leverages this information to recover occluded body parts.
These proxy representations can provide guidance for the neural network and hence make the learning process easier.
% as the proxy representations simplify the observation and are inherently ambiguous in depth and scale.
% However, their learning supervisions are 
However, the proxy representations simplify the observation and introduce additional ambiguity in depth and scale, especially when using proxy representations in a single frame~\cite{song2020human,zhang2020learning, xu2019denserac}.
% Besides, the motion supervision used in the above-mentioned motion capture .
In this work, we alleviate this issue by adopting 2D skeleton sequences as proxy representations and propose to generate proxy data with accurate motions in world space.
% use the simplest proxy representations and learn motion in world space from our proposed proxy-to-motion dataset.

\noindent \textbf{Full-body Motion Capture.} Recent state-of-the-art approaches~\cite{kocabas2021pare,zhang2021pymaf} have achieved promising results for the estimation of body-only~\cite{kocabas2021pare,zhang2021pymaf}, hand-only~\cite{li2022interacting}, and face-only~\cite{DECA_2020} models.
By combining the efforts together, these regression-based approaches have been exploited for monocular full-body motion capture.
% For instance, the pioneering work ExPose
These approaches~\cite{choutas2020monocular,rong2020frankmocap,zhou2021monocular,feng2021collaborative,moon2022Hand4Whole,pymafx2022} typically regress the body, hands, and face models by three expert networks and integrate them together with different strategies.
% of hand and face expert networks are cropped the corresponding image.
% The follow-up works~\cite{rong2020frankmocap,zhou2021monocular,feng2021collaborative,moon2022Hand4Whole} adopt similar strategies and propose several solutions to improve the integration of the hand and body parts.
%%%%%%%%%%%%%%%%new add%%%%%%%%%%%%%%%%%%%
%~\cite{xiang2019monocular,joo2018total} also the full-body capture
%%%%%%%%%%%%%%%%%%%%%%%%%%%%%%%%%%%%%%
For instance, PIXIE~\cite{feng2021collaborative} learns the integration by collaborative regression, while PyMAF-X~\cite{pymafx2022} adopts an adaptive integration strategy with elbow-twist compensation to avoid unnatural wrist poses.
% while maintaining the mesh-to-image alignment.
% suffers from coarse hand alignment
Despite the progress, it remains difficult for existing solutions to run at real-time while being accurate in world space.
% Moreover, these approaches only consider the motion in the image space and produce implausible results in world space.
In this work, we achieve real-time full-body capture with plausible foot-ground contact by introducing new data generation strategies and novel network architectures. 

% \paragraph{Temporal Motion Capture.}
% %%%attention methods
% MPS-Net~\cite{WeiLin2022mpsnet} uses the proposed motion continuity attention module to explore the diverse temporal relations in the motion sequence.

% SPIN~\cite{kolotouros2019learning} initializes an iterative optimization routine with a regression network to fit the SMPL model and supervises the network with the fitted estimate.
% EFT~\cite{joo2021exemplar} proposes an exemplar fine-tuning method to update the network weights of a 3D pose regressor. 

\noindent \textbf{Neural Decent for Motion Capture.} Traditional optimization-based approaches~\cite{bogo2016keep} typically fit 3D parametric models to the 2D evidence but suffer from initialization sensitivity and the failure to handle challenging poses.
To achieve more efficient and robust motion prediction, there are several attempts to leverage the learning power of neural networks for iterative refinement.
HUND~\cite{zanfir2021neural} proposes a learning-to-learn approach based on recurrent networks to regress the updates of the model parameters. 
Song \etal~\cite{song2020human} propose the learned gradient descent to refine the poses of the predicted body model.
Similar refinement strategies are also exploited in PyMAF~\cite{zhang2021pymaf} and LVD~\cite{corona2022learned} by leveraging image features as inputs.
In our work, we propose a contact-aware neural decent module and exploit the usage for more effective motion updates.
% PyMAF~\cite{zhang2021pymaf} leverages a feature pyramid and corrects the predicted parameters based on the mesh-image alignment features.
% LVD~\cite{corona2022learned} iteratively learns the optimal 3D vertex displacement by an ensemble of per-vertex neural fields.       temporal context -> severe depth ambiguity since     + realistic world coordinate system for physical plausible
% Despite their accurate refined results, these methods suffer from severe depth ambiguity due to the inherent absence of depth information in 2D keypoint inputs.

% Physical Plausibility of
\noindent \textbf{Plausible Motion Capture in World Space.} Though existing monocular motion capture methods can produce well-aligned results, they may still suffer from artifacts such as ground penetration and foot skating in world space.
For more physically plausible reconstruction, previous works~\cite{kocabas2021spec,yuan2021glamr} have made attempts to leverage more accurate camera models during the learning process.
To encourage proper contact of human meshes, Rempe \etal~\cite{rempe2020contact} propose a physics-based trajectory optimization to learn the body contact information explicitly.
HuMoR~\cite{rempe2021humor} introduces a conditional VAE to learn a distribution of pose changes in the motion sequence, providing a motion prior for more plausible human pose prediction.
LEMO~\cite{zhang2021learning} learns the proposed motion smoothness prior and optimizes with the physics-inspired contact friction term.
Despite their plausible results, these methods typically require high computation costs and are unsuitable for real-time applications.
For more effective learning of the physical constraints, there are several attempts~\cite{yuan2021simpoe,luo2022embodied} to incorporate the physics simulation in the learning process via reinforcement learning.
However, these methods~\cite{yuan2021simpoe,luo2022embodied} typically depend on 3D scene modeling due to the physics-based formulation.
Recently, there are also attempts to recover camera motion via SLAM techniques~\cite{ye2023slahmr,kocabas2024pace} or regress the human motion trajectory~\cite{rajasegaran2022tracking,sun2023trace}.
Despite the progress, it remains challenging for these methods to run in real-time or produce physically plausible in world space.
In our work, we achieve real-time capture with plausible foot-ground contact in world space by designing novel networks to learn human-centric motion.

%% file: tex/3-data.tex
\section{Proxy Data Generation}

\begin{figure*}[ht!]
    \centering
    \includegraphics[width=\linewidth]{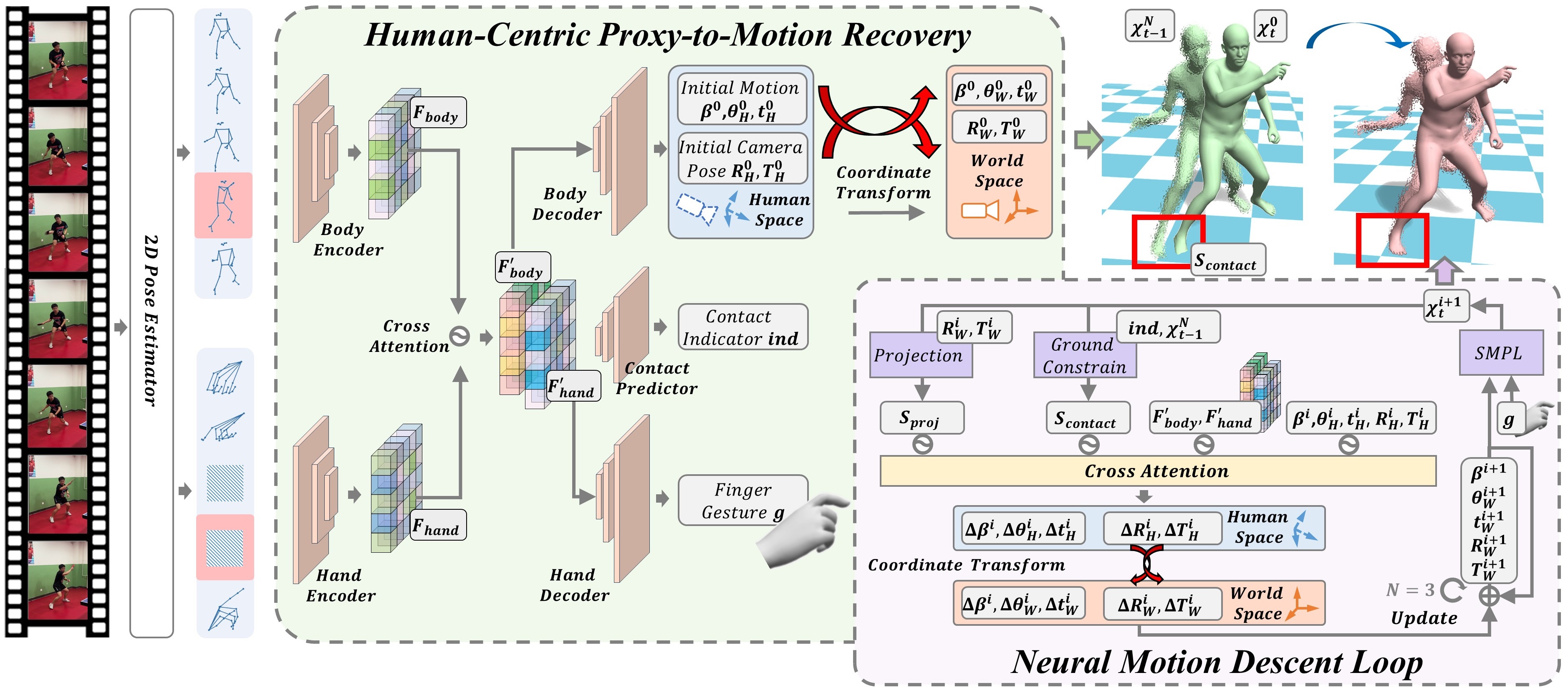}
    \vspace{-8mm}
    \caption{Illustration of the proposed method ProxyCap. 
    Our method takes the estimated 2D skeletons from a sliding window as inputs and estimates the relative 3D motions in the human coordinate space.
    These local movements are accumulated frame by frame to recover the global 3D motions. 
    For more accurate and physically plausible results, a contact-aware neural motion descent module is proposed to refine the initial motion predictions.
    }
    % Specifically, the features of the body and hands are fused by a cross-attention layer and then passed to two branches for body and mesh recovery. 
    \label{fig:overview_pipeline}
    \vspace{-3mm}
\end{figure*}

% To address the precision and generalization challenges associated with extant datasets, 
To tackle the data issue, we synthesize sequential proxy-to-motion data based on 2D skeletons and their corresponding 3D rotational motions in world space.
% During data generation, we also leverage the same motion sequence to synthesize multiple proxy sequences using different camera trajectories, which enhances the learning of the inherent relationship between 2D proxy and 3D motions.
% human movements and postures.
% by synthesizing different camera trajectories over the same motion sequence.
In the following, we describe the synthesis and integration of different types of labels in our proxy data, including body motions, hand gestures, and contact labels.
% propose a simple yet effective strategy to generate accurate motion data in world space.
% We describe the strategy in the following four parts.

\noindent \textbf{Body proxy data.} We adopt the motion sequences from the AMASS dataset~\cite{mahmood2019amass} to generate proxy-to-motion pairs for the body part.
The AMASS dataset is a large-scale body motion sequence dataset that comprises 3,772 minutes of motion sequence data, featuring diverse and complex body poses.
We downsample the motion data to 60 frames per second, resulting in 9407K frames.

\noindent \textbf{Integration with hand gestures.} Since the hand poses in the AMASS dataset are predominantly static, we augment the proxy data with hand poses from the InterHand~\cite{moon2020interhand2} dataset, which contains 1361K frames of gesture data captured at 30 frames per second.
We employ Spherical Linear Interpolation (Slerp) to upsample the hand pose data to 40, 50, and 60 fps and randomly integrate them with the body poses in the AMASS motion sequences.
% to animate the parametric SMPL-X meshes.

\noindent \textbf{Integration with contact labels.} We calculate the continuous contact indicators $ind$ for 3D skeletons as follows:
\begin{equation} \label{equ:contact_indicator}
    \widehat{ind}_i = Sigmoid(\frac{v_{max}-v_i}{k_v})\cdot Sigmoid(\frac{z_{max}-z_i}{k_z})
\end{equation}
where $v_i$ and $z_i$ denote the velocity and the height to the xz-plane of the given joint. $v_{max}$ and $z_{max}$ is set to $0.2m/s$ and $0.08m$, and $k_v$ and $k_z$ is set to $0.04$ and $0.008$.

\noindent \textbf{Camera setting.} For each 3D motion sequence, we generate 2D proxy data under different virtual camera trajectories (four cameras in practice).
Such proxy data enhances the learning of the inherent relationship between 2D proxy and 3D motions and the consistency across different camera viewpoints.
% Therefore our network can learn the consistency of the inherent relative motion differences, which is independent of views.

Specifically, we uniformly sample the field of view (FOV) from $30^{\circ}$ to $90^{\circ}$ for the intrinsic parameters of cameras.
% generate camera trajectories by
When setting the extrinsic parameters for camera trajectories, we position the cameras at distances ranging from $1$ meter to $5$ meters around the human, and at heights varying from $0.5$ meters to $2$ meters to the ground.
Finally, we generate pseudo 2D skeleton annotations by projecting 3D joints into the 2D pixel-plane using these virtual cameras. Moreover, to simulate the jitter of 2D detectors, we add Gaussian noise $\Delta X \sim \mathcal{N}(0, 0.01)$ to 3D joints before projections.

%% file: tex/4-method.tex
\section{Method}
\label{sec:human_motion_recovery}
% \begin{figure*}[ht!]
%     \centering
%     \includegraphics[width=\linewidth]{imgs/human_recovery2.png}
%     \caption{Illustration of human recovery. The upper part (a) depicts the original dilated convolution backbone of ~\cite{videopose3d2020}, while the lower part (b) illustrates our proposed partial dilated convolution architecture. Our approach selectively excludes corrupted data from input signals, allowing us to extract motion features more accurately and effectively. Specifically, detection failure (denoted as green mosaic squares) may occur during extremely fast motion or severe occlusion situations, while our architecture will cut off the connection from corrupted data to prevent disturbance transfer during network forward processing.}
%     \label{fig:human_recovery}
% \end{figure*}

As illustrated in Fig.~\ref{fig:overview_pipeline}, we first detect the 2D skeletons from images and input them into our proxy-to-motion network to recover 3D local motions in human space.
These relative local motions are transformed into a world coordinate system and accumulated along the sliding window. Additionally, we leverage a neural descent module to refine the accuracy and physical plausibility.
In this section, we start with introducing the human-centric setting of our motion prediction.

% In this section, we begin by presenting the camera and motion decoupling approach in Section \ref{subsec:cam_decouple}. Subsequently, we introduce the synthetic data generation process in Section \ref{subsec:synthetic_data_generation}.

\subsection{Human-Centric Motion Modeling}
\label{subsec:cam_decouple}

\begin{figure}[t]
    \centering
    \includegraphics[width=\linewidth]{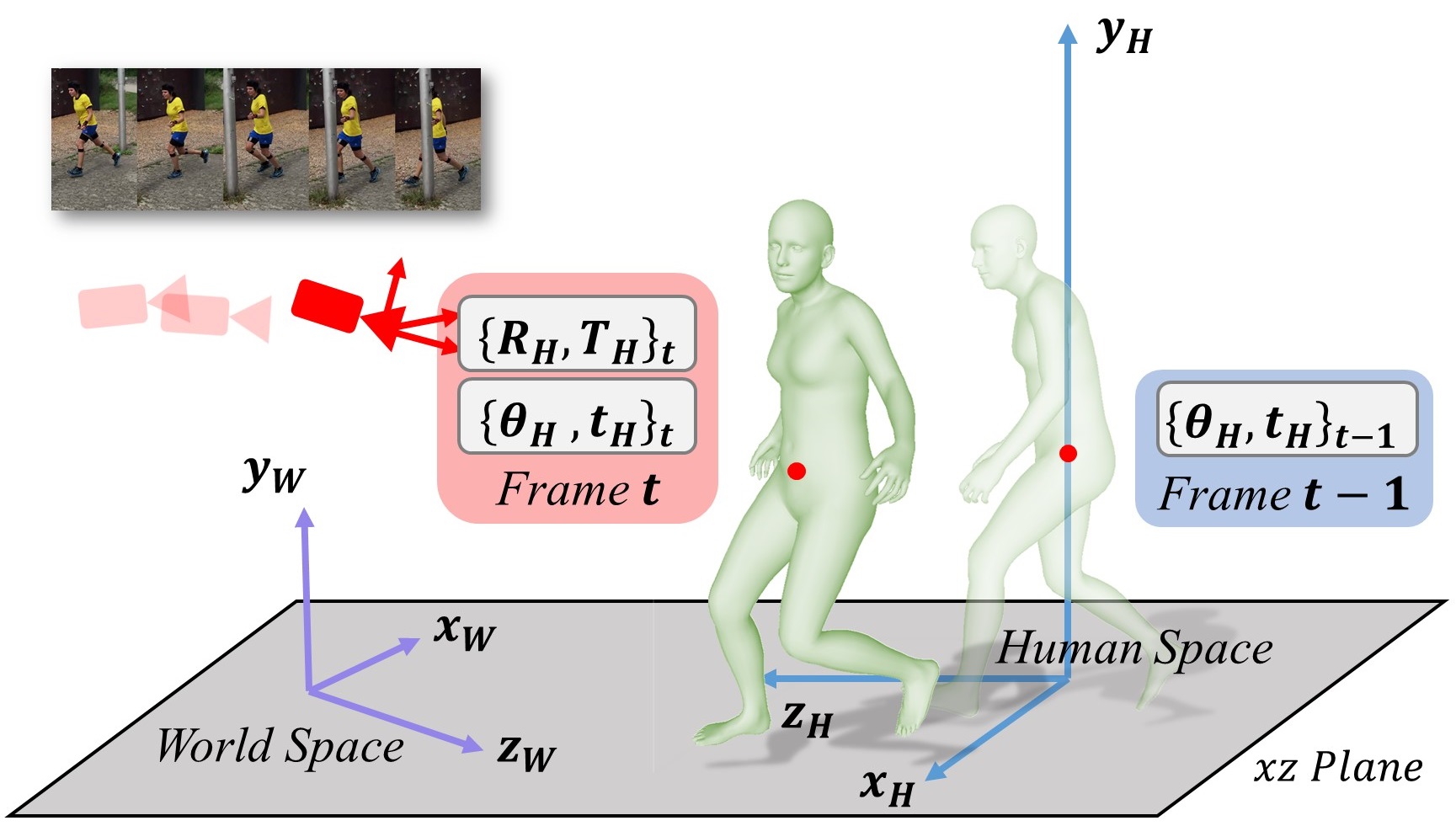}
    % \vspace{-8mm}
    \caption{Illustration of decoupling the world-space motion into the human-centric coordinates and relative camera poses.}
    % \caption{Illustration of the camera coordinate system decoupling. $x_w, y_w, z_w$(blue) denotes the real-world coordinate system, and $x_c, y_c, z_c$(orange) denotes the camera coordinate system. Instead of directly regressing human pose ${\theta}^{c}_b, {t}^{c}_b$ in the camera space, we aim to disentangle them into camera pose($\phi$, $\alpha$, $h$) and the world coordinate space pose ${\theta}^{w}_b, {t}^{w}_b$, which introduces physical priors in training and reference. }
    \label{fig:cam_model}
    \vspace{-5mm}
\end{figure}

For more accurate camera modeling, we adopt the classical pinhole camera model instead of using the simplified orthogonal or weak perspective projection~\cite{videopose3d2020, poseformer2021, PoseTriplet2022,kanazawa2018end, kolotouros2019learning, zhang2021pymaf}.
% we can mitigate the approximation error from camera modeling.
As shown in Fig.~\ref{fig:cam_model}, we transform the global human motion and the camera trajectories into local human space from two adjacent frames, where we adopt $\{\beta\in\mathbf{R^{10}}, \theta \in \mathbf{R^{22\times3}}, t \in \mathbf{R^{3}}\}_{t}$ to denote the parameters of shape, pose, and translation at frame $t$, and $\{R \in \mathbf{R^{3\times 3}}, T \in \mathbf{R^{3}} \}_t$ to denote the camera extrinsic parameters.
Given the pose and shape parameters, the joints and vertices can be obtained within the SMPL Layer: $\{J, V\} =\mathcal{X}\left(\beta, t, \theta, g\right)$.
% where the $g\in \mathbf{R^{2\times15\times3}}$ is the finger gestures.
% We use $\{R \in \mathbf{R^{3\times 3}}, T \in \mathbf{R^{3}} \}_t$ to denote the camera extrinsic parameters.
In the following, we use the subscript \bm{$H$} and \bm{$W$} to distinguish between the human coordinate system and the world coordinate system.

 % modeling of the relative human motion modeling
During the learning of human-centric motions, we adopt a similar setting used in previous works on temporal human motion priors~\cite{ling2020character, rempe2021humor, yuan2021glamr}.
Specifically, each motion sequence will be normalized by a rigid transformation to remove the initial translation in x-z plane and rotate itself with the root orientation heading to the z-axis direction.
With this setting, our network can learn the relative movements of humans, which are independent of observation viewpoints. The detailed implementation of the human-to-world coordinate transformation and the global motion accumulation in world space can be found in our supplementary material.

\subsection{Sequential Full-body Motion Initialization}
\label{sec:temporal_motion_convolution}
As shown in Fig.~\ref{fig:overview_pipeline}, at the first stage of our method, the skeleton sequences are processed by temporal encoders and then fed into decoders for the predictions of the initial motion $\{\mathbf{\beta}^0,\mathbf{\theta}^0,\mathbf{t}^0\}$, initial camera $\{\mathbf{R}^0_{\mathbf{H}},\mathbf{T}^0_{\mathbf{H}}\}$, the contact indicator $\mathbf{ind}$, and the hand poses $\mathbf{g}$.
Following previous baseline~\cite{videopose3d2020}, we build these encoders upon temporal dilated convolution networks.

% % full-convolutional networks with dilated temporal convolutions.
% Moreover, inspired by previous skeleton-based 3D human pose estimation~\cite{Moon_2019_ICCV_3DMPPE, videopose3d2020, poseformer2021, PoseTriplet2022, 2022Ray3D}, we adopt two separate branches in the body-specific regressors to learn the global trajectory and root-relative body motions.
% Specifically, the trajectory branch learns the camera parameters $\alpha, \phi, h$ and the global human transition $t_b$, while the motion branch learns the shape parameter $\beta_b$ and pose parameter $\theta_b$ of the body model.
% Separating these two prediction tasks helps to decouple the global and local motions and hence produce more accurate motions in world space.
% In our implementation, the trajectory branch takes the concatenated $(x, y)$ coordinates of the $J$ joints of $L$ frames as input and outputs the camera pose and global human transition.

For better body-hand compatibility, 
% we leverage both the body and hand context in our network to produce more plausible full-body results.To this end, 
we exploit the cross-attention mechanism to facilitate the motion context sharing during the body and hand recovery.
% the prediction of the body motions and hand gestures.
% More specifically, we exploit a cross-attention layer to fuse the body and hand motion context.
Specifically, we first obtain the initial body features $F_{body}$ and hand features $F_{hand}$ from the temporal encoders and map them as Query, Key, Value matrices in forms of $Q_{body/hand}$, $K_{body/hand}$, and $V_{body/hand}$, respectively.
Then we update the body features $F'_{body}$ and hand features $F’_{hand}$ as follows: 
\begin{equation}
\label{equ:body_hand}
\begin{aligned}
F'_{body} &= V_{body} + Softmax(\frac{Q_{hand} K_{body}^\top}{\sqrt{d_k}}) V_{body},\\
F'_{hand} &=V_{hand} + Softmax(\frac{Q_{body} K_{hand}^\top}{\sqrt{d_k}}) V_{hand}.
\end{aligned}
\end{equation}
% $V'_b=V_b + Softmax(\frac{Q_h K_v^\top}{\sqrt{d_k}}) V_b$ and $V'_h=V_h + Softmax(\frac{Q_b K_h^\top}{\sqrt{d_k}}) V_h$.
% Finally, we feed the full-body motion context, represented as the concatenation of $V'_b$ and $V'_h$, into the contact-aware neural motion descent module to reduce depth ambiguity in optimization.

% As shown in Fig.~\ref{fig:overview_pipeline}, 
The updated features $\{F'_{body}, F'_{hand}\}$ can be further utilized in the contact indicators $ind$ and serve as the temporal context in the Neural Descent module, as will be described shortly.
In our experiments, we demonstrate that the feature fusion in Eq.~\ref{equ:body_hand} can make two tasks benefit each other to produce more comparable wrist poses in the full-body model.

% will then be fed into the corresponding regressors to predict pose parameters.
% With such a fusion strategy, our regressor is able to leverage the body and hand motion context and produce more comparable wrist poses in the full-body model.
\subsection{Contact-aware Neural Motion Descent}
\label{sec:contact_neural_motion_descent}

At the second stage of our method, the initial motion predictions will be refined to be more accurate and physically plausible with the proposed contact-aware neural motion descent module.
% Given the coarse motion predictions, we perform optimization with the contact-aware neural motion descent module.
As shown in Fig.~\ref{fig:overview_pipeline}, this module takes the 2D misalignment and body-ground contact status as input and updates motion parameters during iterations.
% As shown in Fig.~\ref{fig:feedback_net},

\paragraph{Misalignment and Contact Calculation.}
At the iteration of $i\in \{0, 1, ..., N\}$, we calculate the 2D misalignment status by projecting the 3D joints on the image plane and calculate the differences between the re-projected 2D joints and the proxy observations: $\mathcal{S}_{proj} = \Pi(J_i, \{K, R_H^i, T_H^i\}) -\widehat{J}_{2D}$. 
Here, $\Pi(\cdot)$ denotes the perspective projection function, and $K$ denotes the intrinsic parameter.
% which is allowed to be configured from external inputs, or alternatively using the default settings(FOV=${60}^{\circ}$).

For the contact status, we calculate the velocity of 3D joints $v_{xz}^i$ in xz-plane and the distance to the ground as $d_{y}^i$.
Moreover, we also leverage the temporal features from inputs 2D skeletons to predict the contact labels $ind$, which will be used as an indicator to mask the body-ground contact.
Then, the contact status of the current predictions can be obtained as $\mathcal{S}_{contact} = ind \odot (v_{xz}^i, d^i)$, where $\odot$ denotes the Hadamard product operation.

\begin{figure}[t]
    \centering
    \includegraphics[width=\linewidth]{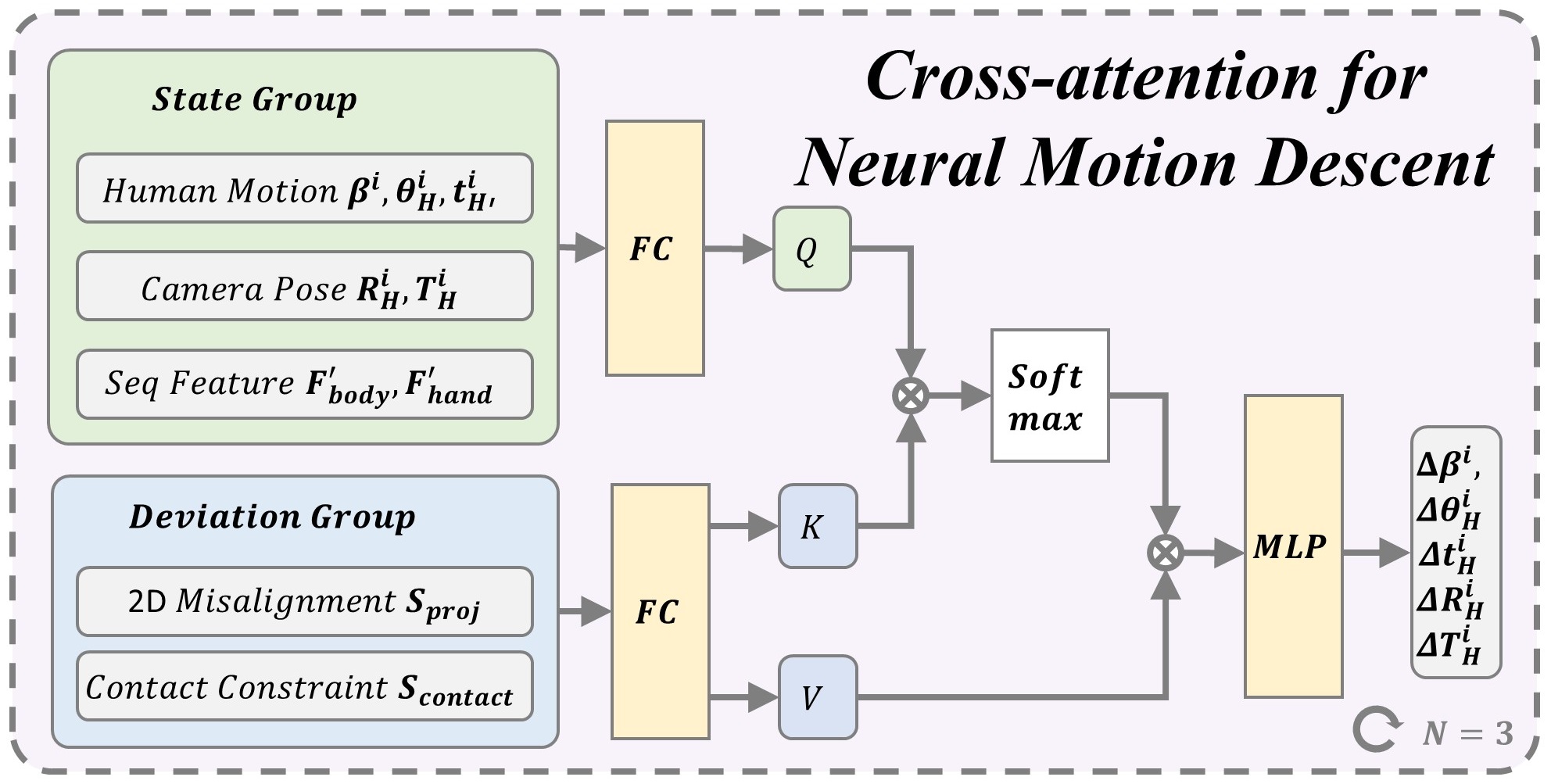}
    \vspace{-8mm}
    \caption{Implementations of the neural descent module.
    % With the consideration of the modality differences between state group and deviation group, we leverage an effective cross-attention scheme to learn the update motions.
    % We initially transform the estimated SMPL parameters $\theta^w, t^w$ to the reference space as $\theta^{ref}, t^{ref}$, and subsequently feed them to the contact-aware neural motion descent module. 
    % Furthermore, the encoder is enhanced via a cross-attention approach that rectifies the parameter space of input tensors to significantly boost the performance.
    }
    \vspace{-6mm}
    \label{fig:feedback_net}
\end{figure}

\paragraph{Motion Update.}
After obtaining the contact and misalignment status, we feed them into the neural motion descent module for motion updates.
As shown in Fig.~\ref{fig:feedback_net}, the descent module takes the two groups of tensors as input: 
i) the state group includes the current SMPL parameters in the human coordinate system $\beta^i, t^i_H, \theta^i_H$, camera pose $R^i_H, T^i_H$ and the sequential motion context $F_{seq}=\{F'_{body}, F'_{hand}\}$;
ii) the deviation group includes the current misalignment status $\mathcal{S}_{proj}$ and contact status $\mathcal{S}_{contact}$.

A straightforward solution would be using an MLP to process these two groups of tensors.
However, the values of these two groups exhibit significant differences.
For instance, the values of the state tensors change smoothly while the values of the deviation tensors may change rapidly along with the misalignment and contact status.
Simply concatenating them as inputs introduces difficulty in the learning process.
Note that the magnitude of the deviation tensors is highly correlated with the parameter updates.
When the body model is well-aligned without foot skating or ground penetration, the values of the deviation tensors are almost zero, indicating that the refiner should output zeros to prevent further changes in the pose parameters.
Otherwise, the refiner should output larger update values for motion adjustments.
To leverage such a correlation property, we exploit a cross-attention module to build a more effective architecture.

As shown in Fig.~\ref{fig:feedback_net}, two fully connected layers are leveraged to process the tensors of the state and deviation groups and produce the Query, Key, and Value for the cross-attention module.
In this way, our contact-aware neural motion descent module can effectively learn the relationship between the state and deviation groups and hence produce more accurate motion updates.
Moreover, the sequential motion context $F_{seq}$ is also leveraged in our neural descent module to mitigate the depth uncertainty and improve the motion predictions.
% For instance, the magnitude of residual tensors is highly correlated with learned parameter updates.
% When SMPL meshes are well-aligned without foot skating or ground penetration, the values of the misalignment status are almost zero, indicating that the network should output zeros to prevent further changes to the SMPL parameters.
% When SMPL meshes are well-aligned without foot skating or ground penetration, the resulting residual tensors 
% $\mathcal{L}{proj}$ and $\mathcal{L}{contact}$
% are almost zero, indicating that the network should output zeros to prevent further changes to the SMPL parameters.
% In contrast, the state tensors require only minor adjustments during the iteration process.
% Therefore, we introduce a cross-attention architecture to effectively fuses them while ensuring the magnitude correlation between residuals and updates, which is illustrated in Fig.~\ref{fig:feedback_net}.
% In addition, the sequential context in the module contains time constraints in motion sequence, which mitigate depth uncertainty and significantly improve the performance of neural descent.

Compared with previous work~\cite{rempe2021humor, zhang2021learning,song2020human, zanfir2021neural}, the proposed contact-aware neural motion descent module offers the advantage of freeing us from the heavy cost of explicit gradient calculations or the manual tuning of hyper-parameters during testing.
Furthermore, the module is capable of learning human motion priors with contact information from our synthetic dataset, which provides a more suitable descent direction and steps to escape the local minima and achieve faster convergence.

\subsection{Loss Function}
\label{sec:loss_function}
In our solution, the full-body motion recovery module and the contact-aware neural motion descent module are trained sequentially.
Benefiting from the proxy-to-motion learning, the ground-truth full-body pose $\theta, g$, and human body shape $\beta_b$ can be obtained for supervision from our synthetic dataset.
Overall, the objective function of motion recovery can be written as follows:
\begin{equation} \label{equ:loss_rec}
\mathcal{L}_{rec} = \mathcal{L}_{3D} + \mathcal{L}_{2D} + \mathcal{L}_{\theta} + \mathcal{L}_{\beta}+ \mathcal{L}_{cam} + 
\mathcal{L}_{consist} + \mathcal{L}_{smooth} 
\end{equation}

Specifically, $\mathcal{L}_{3D}$ involves 3D MPJPE loss and 3D trajectory L1 loss while $\mathcal{L}_{2D}$ is the projected 2D MPJPE loss.
$\mathcal{L}_{\theta}, \mathcal{L}_{\beta}, \mathcal{L}_{cam}$ represents L1 loss between the estimated human pose, shape and camera pose to our synthetic ground truth.
$\mathcal{L}_{consist}$ is a L1 loss to restrict the consistency of the local motion outputs $\theta_{H}, t_{H}$ of the same 3D motion sequence via different observations from virtual cameras.
$\mathcal{L}_{smooth}$ is adopted from ~\cite{zeng2021smoothnet} by penalizing the velocity and acceleration between the estimation and the ground truth.
For the neural descent module, the objective loss can be written as:
\begin{equation} \label{equ:loss_descent}
\begin{cases}
     \mathcal{L}_{desc} = \sum_k {u^{N-k} (\mathcal{L}_{rec} + \mathcal{L}_{contact})} \\
     \mathcal{L}_{contact} =\sum_i{ind^{gt}\odot(||v_{xz}||_2 + ||d_y||_2)} \\
     \mathcal{L}_{ind} = \sum_i{Entropy(ind^{gt}, ind^{est})}
\end{cases}
\end{equation}
where $k=1,2,..., N$ is the iteration time and $u$ is the decay ratio to emphasize the last iteration.
% where $k=1,2,..., K$ is the iteration time and $u$ is the decay ratio to balance each iteration.
We set $K=3, u=0.8$ in practice.
$\mathcal{L}_{contact}$ involves the error of trajectory drifting, foot floating, or ground penetration. $\mathcal{L}_{ind}$ refers to the loss between the predicted contact label to the ground truth.

%% file: tex/5-experiments.tex
\section{Experiments}

In this Section, we validate the efficacy of our method and demonstrate accurate human motion capture results with physically plausible foot-ground contact in world space.
 % As shown in Fig~\ref{fig:results},

\noindent \textbf{Dataset.} The RICH dataset \cite{huang2022capturing} is collected with a multi-view static camera system and one moving camera that the ground truth 3D human motions can be recovered using spatial stereo geometry.
The EgoBody~\cite{Zhang:ECCV:2022} is captured by a multi-camera rig and a head-mounted device, focusing on the social interactions in 3D scenes.
Dynamic Human3.6M is a benchmark proposed in ~\cite{yuan2021glamr} to simulate the moving cameras on Human3.6M~\cite{ionescu2014human3} by randomly cropping with a small view window around the person.

\noindent \textbf{Metrics.} In our experiments, we follow previous work~\cite{yuan2021glamr} to report various metrics, primarily focusing on the evaluation of the world coordinate system. The WA-MPJPE metric reports the MPJPE after aligning the entire trajectory of both the predicted and GT through Procrustes Alignment. The W-MPJPE metric reports the MPJPE after aligning the first frame of sequences. The PA-MPJPE metric reports the MPJPE error after applying the ground truth trajectories to each frame. The ACCEL metric is used to evaluate the joint acceleration.

\begin{figure*}[ht!]
    \centering
    \includegraphics[width=\linewidth]{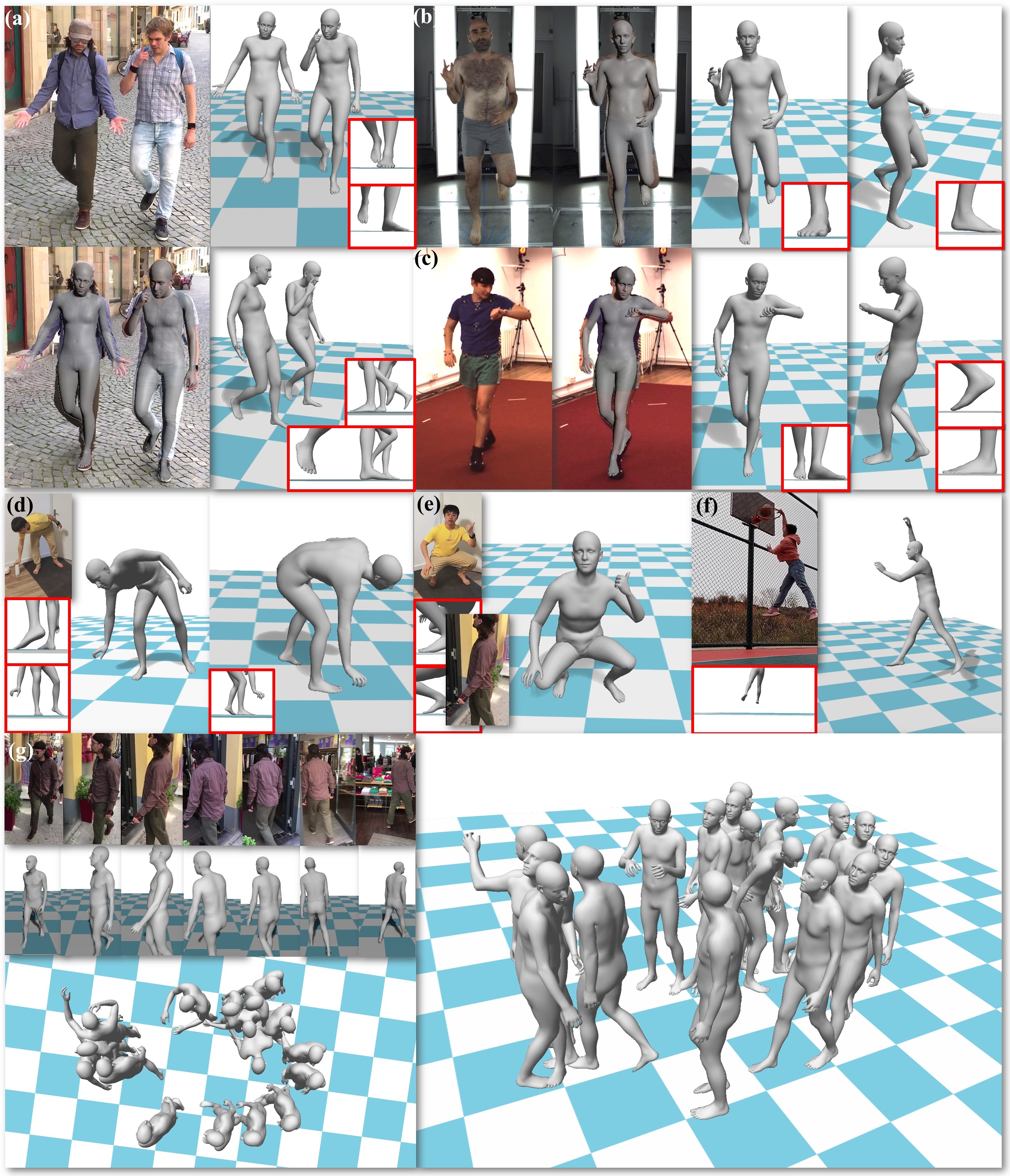}
    \caption{Results across different cases in the (a,g) 3DPW~\cite{vonMarcard2018}, (b) EHF~\cite{SMPL-X:2019}, and (c) Human3.6M~\cite{ionescu2014human3}  datasets and (d,e,f) internet videos. We demonstrate that our method can recover the accurate and plausible human motions in moving cameras at a real-time performance. Specifically, (g) demonstrates the robustness and the temporal coherence of our method even under the occlusion inputs. }
    \label{fig:results}
    \vspace{-3mm}
\end{figure*}

\subsection{Comparison with the State of the Art}

% \begin{table}[t]
% \centering
% \caption{Quantitative comparison on the EgoBody~\cite{Zhang:ECCV:2022} dataset. The symbol $\dag$ denotes the methods relying on SLAM.}
% \vspace{-3mm}
% \resizebox{\linewidth}{!}{
% \begin{tabular}{lcccc}
% \hline
% Methods & W-MPJPE $\downarrow$ & WA-MPJPE $\downarrow$ & PA-MPJPE $\downarrow$ & ACCEL $\downarrow$ \\ \hline
% $\dag$ SLAHMR~\cite{ye2023slahmr}  & \textbf{141.1 }& 101.2 & 79.13 & 25.78 \\
% $\dag$ PACE~\cite{kocabas2024pace} & 147.9 & \textbf{101.0} & \textbf{66.5} & \textbf{6.7}\\ 
% \hline
% GLAMR~\cite{yuan2021glamr} & 416.1 & 239.0 & 114.3 & 173.5 \\
% Ours  & \textbf{385.5} & \textbf{131.3} & \textbf{73.5} &  \textbf{49.6}\\ \hline
% \end{tabular}}
% \label{tab:egobody}
% \end{table}

% \begin{table}[t]
% \centering
% \caption{Quantitative comparison on RICH~\cite{huang2022capturing}. The symbol $\dag$ denotes the methods relying on SLAM.}
% \vspace{-3mm}
% \resizebox{\linewidth}{!}{
% \begin{tabular}{lccccc}
% \hline
% Methods & W-MPJPE $\downarrow$ & WA-MPJPE $\downarrow$ & PA-MPJPE $\downarrow$ & ACCEL $\downarrow$ \\ \hline
% $\dag$SLAHMR~\cite{ye2023slahmr} & 571.6 & 323.7  & 52.5 & 9.4 \\
% $\dag$PACE~\cite{kocabas2024pace} & \textbf{380.0} & \textbf{197.2}  & \textbf{49.3} & \textbf{8.8} \\ 
% \hline
% GLAMR~\cite{yuan2021glamr} & 653.7 & 365.1  & 79.9 & 107.7 \\
% Ours & \textbf{629.8} & \textbf{343.6} & \textbf{56.0} & \textbf{25.3} \\ 
% \hline
% \end{tabular}}
% \label{tab:rich}
% \end{table}

\begin{table}[t]
\centering
\caption{Quantitative comparison on EgoBody~\cite{Zhang:ECCV:2022} and RICH~\cite{huang2022capturing}. The symbol $\dag$ denotes the methods relying on SLAM.}
% \vspace{-3mm}
\resizebox{\linewidth}{!}{
\begin{tabular}{lcccc}
\hline
Methods & W-MPJPE $\downarrow$ & WA-MPJPE $\downarrow$ & PA-MPJPE $\downarrow$ & ACCEL $\downarrow$ \\ \hline
\textbf{EgoBody dataset}  &  &   &   &   \\
$\dag$ SLAHMR~\cite{ye2023slahmr}  & \textbf{141.1 }& 101.2 & 79.13 & 25.78 \\
$\dag$ PACE~\cite{kocabas2024pace} & 147.9 & \textbf{101.0} & \textbf{66.5} & \textbf{6.7}\\ 
% \hline
GLAMR~\cite{yuan2021glamr} & 416.1 & 239.0 & 114.3 & 173.5 \\
Ours  & \textbf{385.5} & \textbf{131.3} & \textbf{73.5} &  \textbf{49.6}\\ \hline
\textbf{RICH dataset}  &  &   &   &   \\
$\dag$SLAHMR~\cite{ye2023slahmr} & 571.6 & 323.7  & 52.5 & 9.4 \\
$\dag$PACE~\cite{kocabas2024pace} & \textbf{380.0} & \textbf{197.2}  & \textbf{49.3} & \textbf{8.8} \\ 
% \hline
GLAMR~\cite{yuan2021glamr} & 653.7 & 365.1  & 79.9 & 107.7 \\
Ours & \textbf{629.8} & \textbf{343.6} & \textbf{56.0} & \textbf{25.3} \\ 
\hline
\end{tabular}}
\label{tab:egobody_rich}
\vspace{-4mm}
\end{table}

We compare our approach with the state-of-the-art approaches to human motion recovery under dynamic cameras, including GLAMR~\cite{yuan2021glamr}, SLAMHR~\cite{ye2023slahmr} and PACE~\cite{kocabas2024pace}. 
Both the SLAMHR and PACE require a pre-processing SLAM to reconstruct the scene to solve the camera trajectories (refer to Tab.~\ref{tab:performance}).
Such a process is time consuming and requires texture-rich backgrounds, which narrows their applications. 
To validate the effectiveness of the proposed solution, we primarily compare our method with GLAMR, as it also runs without SLAM.
 % our method is 
We also conduct comparison experiments on the RICH and EgoBody datasets, as shown in Tab.~\ref{tab:egobody_rich}.
% Tab.~\ref{tab:egobody} and Tab.~\ref{tab:rich}.
As shown in the table, our method achieves significant improvements over previous solutions in all metrics.
% through our proxy-to-motion learning scheme and the neural descent module.

\begin{figure}[ht!]
    \centering
    \includegraphics[width=\linewidth]{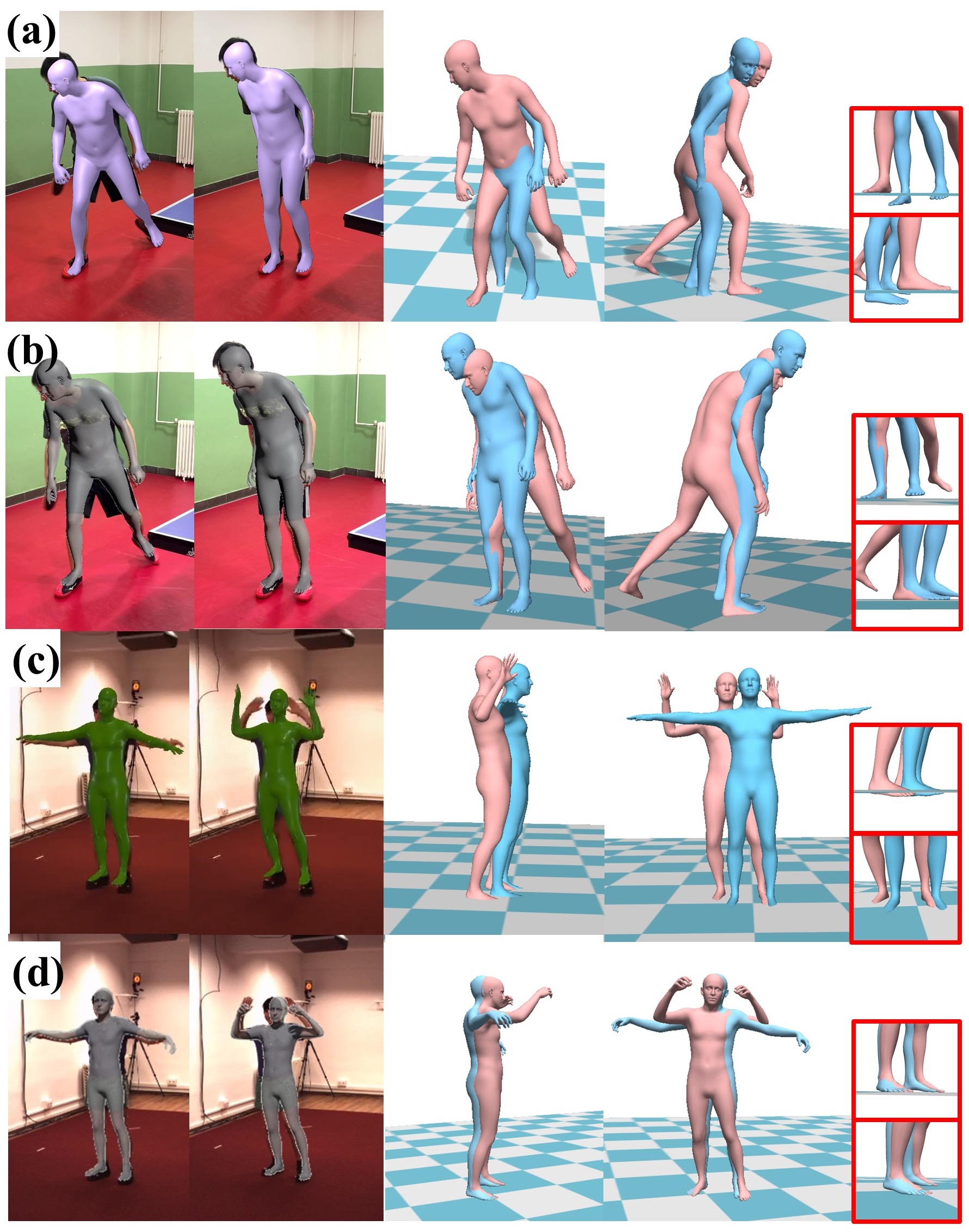}
    \caption{Qualitative comparison with previous state-of-the-art methods: (a) PyMAF-X~\cite{pymafx2022}, (c) GLAMR~\cite{yuan2021glamr}, (b)(d) Ours.}
    \label{fig:compare_glamr}
    \vspace{-3mm}
\end{figure}

Visual comparisons with previous different solutions are also depicted in Fig.~\ref{fig:compare_glamr} and the video in our supplementary materials, where our method again shows superior results in terms of model-image alignment and foot-ground contact in world space.
% Moreover, we also compares visually in Fig.~\ref{fig:compare_glamr} and the supplementary video.

% \begin{figure}[ht!]
%     \centering
%     \includegraphics[width=\linewidth]{imgs/compare_cvpr.jpg}
%     \caption{Qualitative comparison with previous state-of-the-art methods: LearnedGD~\cite{song2020human}, PyMAF-X~\cite{pymafx2022}, and GLAMR~\cite{yuan2021glamr}.}
%     \label{fig:compare_glamr}
%     \vspace{-3mm}
% \end{figure}

\subsection{Ablation Studies}
We conduct ablation studies to validate the effectiveness of the proposed neural descent method on the Dynamic Human3.6M dataset following the setup of ~\cite{yuan2021glamr}. As shown in Tab~\ref{tab:ablation}, the Neural Descent module can significantly reduce the motion errors in world space.

\begin{table}[ht]
\centering
\caption{Ablation study of the Neural Descent module on the Dynamic Huaman3.6M dataset.}
% \vspace{-3mm}
\begin{tabular}{ccc}
\hline
Neural Descent & W-MPJPE $\downarrow$  & PA-MPJPE $\downarrow$ \\  \hline
w/o  & 644.8 & 48.4  \\
w/   & \textbf{605.4} & \textbf{45.9}  \\
\hline
\end{tabular}
\label{tab:ablation}
\end{table}

\begin{figure}[ht!]
    \centering
    \includegraphics[width=0.8\linewidth]{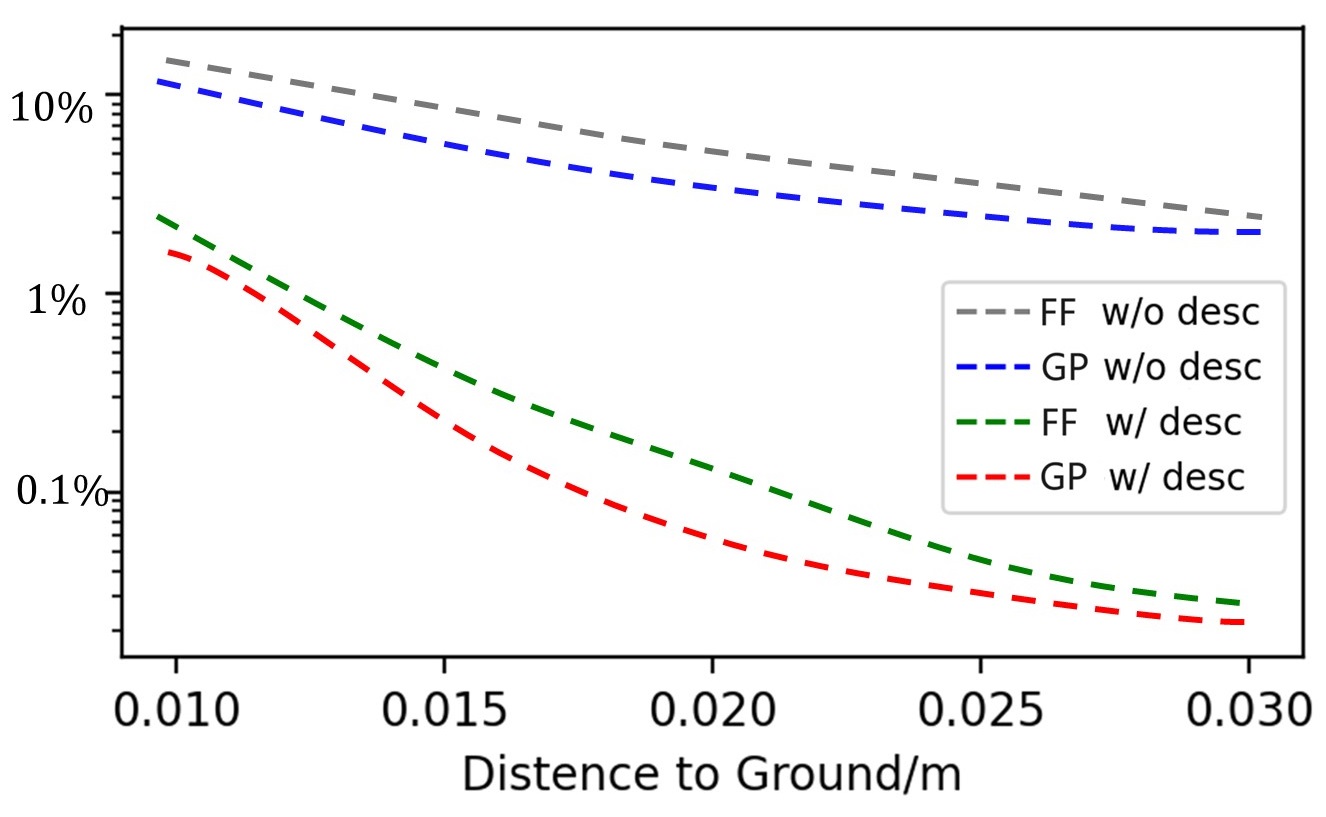}
    \vspace{-3mm}
    \caption{The ablation study on the percentage of foot floating (FF) and ground penetration (GP). We vary the threshold from 1cm to 3cm to calculate the corresponding FF and GP metrics.}
    \vspace{-2mm}
    \label{fig:contact}
\end{figure}

We also report the metric of ground penetration~\cite{yuan2021simpoe} (GP) and foot floating (FF) in the Human3.6M~\cite{ionescu2014human3} dataset. 
The GP is defined as the percentage of the frames that penetrate to the ground. The FF is defined as the percentage of frames with foot-ground distances far from the given threshold.
We report the curves of GP and FF with respect to the distance to ground in Fig.~\ref{fig:contact} with a logarithmic scale, where we can conclude that the neural descent algorithm can significantly improve the ground contact plausibility.

\subsection{Runtime}
It is also worth noting that the proposed method has improved the speed by an order of magnitude compared to the previous methods. 
The speeds of different methods are reported in Tab.~\ref{tab:performance}. 
Our method can reach real-time performance at 30 FPS in a laptop with RTX4060, which is very promising to enable various applications related to virtual humans.

\begin{table}[ht]
\centering
\caption{Runtime comparison with the state-of-the-art methods.}
% \vspace{-3mm}
\begin{tabular}{ccccc}
\hline
Method & SLAMHR  & PACE & GLAMR & Ours \\ \hline 
FPS  & 0.04 & 2.1 & 2.4 & \textbf{30}  \\
\hline
\end{tabular}
\label{tab:performance}
\vspace{-3mm}
\end{table}

%% file: tex/6-conclusion.tex
\section{Conclusion}
In this paper, we present ProxyCap, a real-time monocular full-body motion capture approach with physically plausible foot-ground contact in world space.
We leverage a proxy dataset based on 2D skeleton sequences with accurate 3D rotational motions in world space.
Based on the proxy data, our network learns motions from a human-centric perceptive, which enhances the consistency of human motion predictions under different camera trajectories.
For more accurate and physically plausible motion capture, we further propose a contact-aware neural motion descent module so that our network can be aware of foot-ground contact and motion misalignment.
Based on the proposed solution, we demonstrate a real-time monocular full-body capture system under moving cameras.

\paragraph{Limitations.}
As our method recovers 3D motion from 2D joint observations, the depth ambiguity issue remains especially when the person is captured in nearly static poses.

% To human-centric proxy-to-motion learning, 

% leverage a human-centric proxy-to-motion learning scheme and 
% Besides, we propose a contact-aware neural motion descent module for more accurate and physically plausible results.

%
% Besides, we leverage the body-hand context information in our network so that the wrist poses can be more compatible with the full-body model.
%
% We demonstrate a real-time monocular full-body capture system and show superior results to existing solutions with more plausible foot-ground contact.

% \paragraph{Limitations.}

% Similar to previous work~\cite{song2020human}, it is difficult for our method to capture the body shape from sparse skeletons.
% Meanwhile, though our method shows more plausible results in world space, the mesh-to-image alignment is slightly inferior to state-of-the-art image-based solutions~\cite{zhang2021pymaf,pymafx2022} especially when the camera setting differs from the training data.

%% file: tex/7-supp.tex
\section{Implementation Details}
\subsection{Human-to-World Coordinate Transformation}
We estimate the local human pose of each frame in our proxy-to-motion network, then we transform it into the global world space. In the first frame, the accumulated translation of human in x-z plane $t_{xz}$ is set to zero. For the later human-space estimations, we firstly rotate the front axis of camera in human space to align the y-z plane by $R_{front}$. Denote the target parameters of camera in world space as $R_W, T_W$ and the predicted parameters in human space as $R_H, T_H$. We have $R_W = R_H \cdot R_{front}$, $T_W = -R_W \cdot (R_{front}^T \cdot (-R_H^T \cdot T_H) + t_{xz})$.
And the human orientation should also be rotate in the same time to maintain relative stillness: $\theta_W(root) = R_{front}^T \cdot \theta_H(root)$. The world translation can be calculated by $t_W = R_{front}^T \cdot (t_H+J_{root})-J_{root} + t_{xz}$. Here $J_{root}$ is the root joint of SMPL model in T-pose to eliminate the error resulted from the root-misalignment with the original point of SMPL model. Finally, the accumulated translation of human in x-z plane $t_{xz}$ is updated by $t_{xz}^{t+1} = t_{xz}^{t} + R_{front}^T \cdot t_H$.

\subsection{Partial Dilated Convolution}
\begin{figure*}[ht!]
    \centering
    \includegraphics[width=\linewidth]{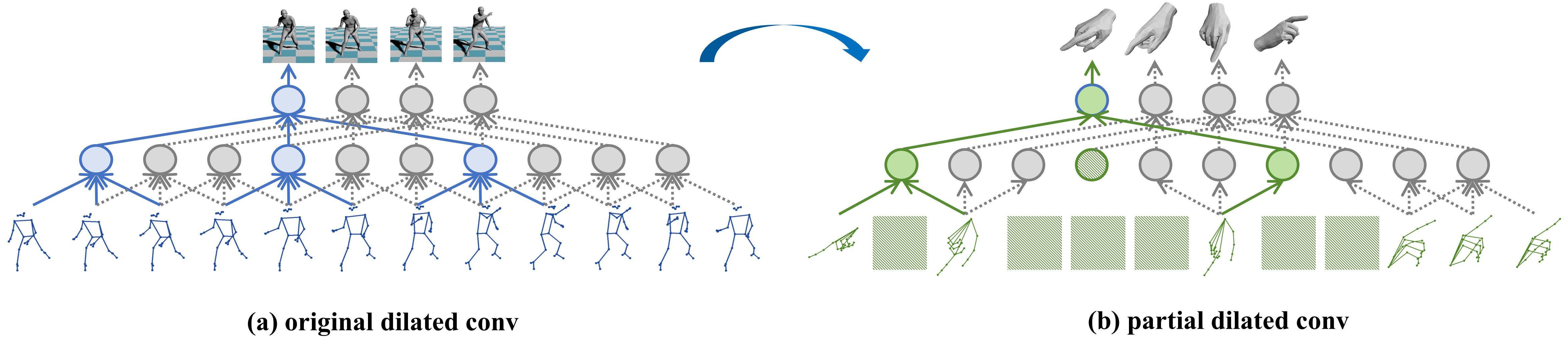}
    \caption{Illustration of human recovery. The left part (a) depicts the original dilated convolution backbone of ~\cite{videopose3d2020}, while the right part (b) illustrates our proposed partial dilated convolution architecture. Our approach selectively excludes corrupted data from input signals, allowing us to extract motion features more accurately and effectively. Specifically, detection failure (denoted as green mosaic squares) may occur during extremely fast motion or severe occlusion situations, while our architecture will cut off the connection from corrupted data to prevent disturbance transfer during network forward processing.}
    \label{fig:human_recovery}
\end{figure*}

It should be noted that the hand area, being an extremity, is more prone to being affected by heavy motion blur and severe occlusions, resulting in missing detections.
Simply setting the corrupted data to zero is not a viable solution as the original convolution kernel is unable to distinguish between normal data and corrupted data, leading to a significant reduction in performance as noise is propagated through the network layers.

To overcome this challenge, we employ partial convolution~\cite{partialconv2018} to enhance our 1D dilated convolution framework.
As illustrated in Fig.~\ref{fig:human_recovery}, rather than indiscriminately processing the input signals as in the original convolution operator, we utilize a mask-weighted partial convolution to selectively exclude corrupted data from the inputs.
This enhances the robustness of hand recovery in scenarios involving fast movement and heavy occlusion.
Specifically, the latent code $X_0$ is initially set as the concatenation of the $(x, y)$ coordinates of the $J$ joints for each frame $f\in1, 2, ..., L$,  while the mask $M_0$ is initialized as a binary variable with values of 0 or 1, corresponding to the detection confidence. Then we integrated the 2D partial convolution operation and mask update function from ~\cite{partialconv2018} into our 1D dilated convolution framework:
\begin{equation} \label{equ:partial_conv}
    \begin{cases} 
    M_{k+1} = I\left(sum(M_k)>0\right) \\
    X_{k+1} = M_{k+1} \left( W_k^\top \left(X_k \odot M_k \right) \frac{size(M_k)}{sum(M_k)} + b_k \right)
    \end{cases}
\end{equation}
where $W$ and $b$ denotes the convolution filter weights and bias at layer $k$, and $\odot$ denotes the element-wise multiplication. Furthermore, in the training procedure, half of the sequential inputs are randomly masked to simulate detection failures that may occur in a deployment environment.

\subsection{Training}
We train our network using the Adam optimizer with a learning rate of 1e-4 and a batch size of 1024 in NVIDIA RTX 4090.
We adopt a two-stage training scheme. Firstly, we train our proxy-to-motion initialization network (Sec.~4.2) for 50 epochs.
Subsequently, we fix the weights of the motion recovery network and train the neural descent network (Sec.~4.3) for another 50 epochs. 

\subsection{Proxy Dataset}
We conduct the training process on our synthetic proxy dataset. The 3D body rotational motions of the training set are sampled from AMASS~\cite{mahmood2019amass}: [ACCAD, BMLmovi, BMLrub, CMU, CNRS, DFaust, EKUT, Eyes Japan Dataset, GRAB, HDM05, HumanEva, KIT, MoSh, PosePrior, SFU, SOMA, TCDHands, TotalCapture, WEIZMANN] and [SSM, Transitions] are for testing. Otherwise, for the generation of hand motions,  we adopt the same dataset division of  InterHand~\cite{moon2020interhand2}. Then We animate the SMPL-X mesh and generate virtual cameras to obtain the pseudo 2D labels.

\section{Computational Complexity}
In this section, we compare the inference speed of our method.
Our real-time monocular full-body capture system can be implemented on a single Laptop (NVIDIA RTX 4060 GPU).
Specifically, for the 2D pose estimator, we leverage the off-the-shelf Mediapipe~\cite{lugaresi2019mediapipe} and MMPose and re-implement it on the NVIDIA TensorRT platform.
We report the inference time of each module in Table.~\ref{tab:performance}. 

\begin{table}[ht!]
\caption{Time costs of each module in our pipeline.}
\centering
\begin{tabular}{ccc}
\hline
Network & Input &Speed \\ \hline
Body Crop Net &$224\times224\times3$ &  2ms \\
Body Landmark Net &$384\times288\times3$  & 5ms \\ 
Hand Crop Net &$2\times256\times256\times3$  & 1.5ms \\
Hand Landmark Net &$256\times256\times3$  & 2ms \\ 
Pose Initialization Net &$81\times67\times2$  & 3ms \\ 
Neural Descent Net & $\backslash$ & 10ms \\ \hline
\end{tabular}
% \vspace{2mm}
\vspace{-3mm}
\label{tab:performance}
\end{table}